\documentclass[10pt, a4paper]{article}
\usepackage{float}

\usepackage{lrec-coling2024} 
\usepackage{tabularray}
\usepackage{soul}
\usepackage{placeins}
\usepackage{tcolorbox}
\usepackage{caption}
\usepackage{comment}
\usepackage{multirow}
\usepackage{graphicx}
\usepackage{authblk}
\usepackage{kotex}
\usepackage{amsmath}
\usepackage{subcaption}

\usepackage{todonotes} 

\usepackage{booktabs} 

\title{K-pop Lyric Translation: Dataset, Analysis, and Neural-Modelling}

\name{Haven Kim $^1$ Jongmin Jung $^2$ Dasaem Jeong $^3$ Juhan Nam $^1$} 

\address{$^1$ Graduate School of Culture Technology, KAIST \\
$^2$ Department of Artificial Intelligence, Sogang University \\
$^3$ Department of Art and Technology, Sogang University\\
         khaven@kaist.ac.kr, jongmin@sogang.ac.kr, dasaemj@sogang.ac.kr, juhan.nam@kaist.ac.kr
}

\abstract{
Lyric translation, a field studied for over a century, is now attracting computational linguistics researchers. We identified two limitations in previous studies. Firstly, lyric translation studies have predominantly focused on Western genres and languages, with no previous study centering on K-pop despite its popularity. Second, the field of lyric translation suffers from a lack of publicly available datasets; to the best of our knowledge, no such dataset exists. To broaden the scope of genres and languages in lyric translation studies, we introduce a novel singable lyric translation dataset, approximately 89\% of which consists of K-pop song lyrics. This dataset aligns Korean and English lyrics line-by-line and section-by-section. We leveraged this dataset to unveil unique characteristics of K-pop lyric translation, distinguishing it from other extensively studied genres, and to construct a neural lyric translation model, thereby underscoring the importance of a dedicated dataset for singable lyric translations.
 \\ \newline \Keywords{Lyric Translation, K-pop Translation, Lyrics Information Processing} }

\begin{document}

\maketitleabstract

\section{Introduction}
Singable lyric translation is a common practice to bolster the global resonance and appeal of music across diverse genres, from opera and animated musical songs (such as those from Disney) to children's songs and hymns~\cite{mateo2012music}. With the continuous globalization of music, the importance and popularity of singable lyric translation are increasing~\cite{susam2008translation}, particularly on social media platforms like YouTube.

Despite its widespread appeal, singable lyric translation is acknowledged as a challenging discipline as it calls for a sufficient understanding of musicology and linguistics~\cite{Spaeth1915, susam2008translation}. This challenge was underlined as far back as the 19th century by Richard Wagner, who criticized the opera translations he encountered, describing them as sounding like textbooks because, in his view, the translators lacked musical knowledge~\cite{wagner}. Moreover, previous research emphasizes that lyric translation also requires solid cultural consideration, given the distinct poetic norms of each language~\cite{low2008, cultural2008, case2009}. Consequently, due to these inherent intricacies, the study of lyric translation remains largely unexplored. While some studies have endeavored to investigate singable lyric translations, their research has primarily centered on Western languages, predominantly English and German, and Western genres, such as opera and animated musical songs~\cite{german_comedy2005, theatre_opera_2007, low2008, theatre2017,  disney2019}. To our knowledge, there has been no comprehensive analysis of Korean pop (K-pop) translations, despite their substantial popularity on social platforms.

\begin{figure}[t]
 \centerline{
 \includegraphics[width=0.75\linewidth]{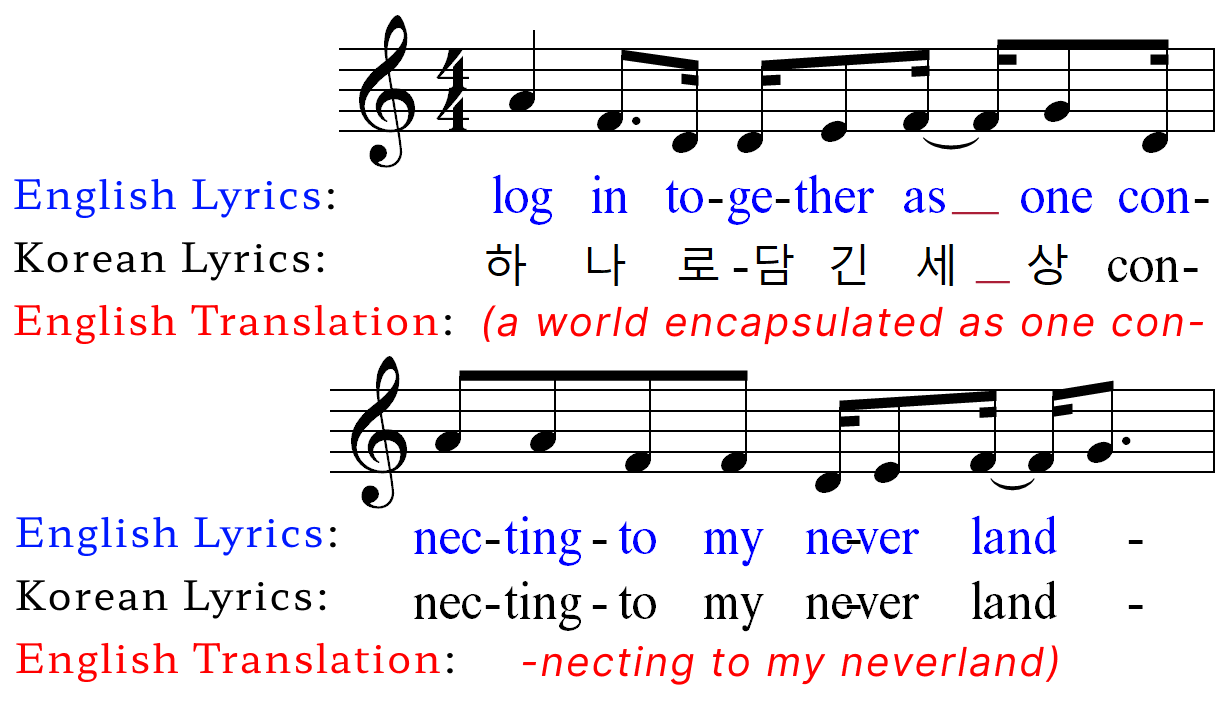}}
\caption{An illustration of K-pop translation, featuring ``ID Peace B'' by BoA, with English singable lyrics, Korean singable lyrics, and their corresponding non-singable English translations.}
 \label{fig:id_peace_b}
\end{figure}

Another challenge in lyric translation studies is the absence of a publicly available dataset. As far as we can tell, no public singable lyric translation dataset currently exists, creating a barrier to fully deciphering the art of lyric translation. Hence, systematic analysis of singable lyric translation has primarily relied on individual case studies \cite{german_comedy2005, case2009, case_musical2010, disney2019}. Moreover, while automatic lyric translation has gained popularity, the development of neural lyric translation models has been largely dependent on semi-supervised methods~\cite{acl2022, acl2023} or privately sourced datasets \cite{li2023translate}.

 \begin{figure*}[t]
 \centerline{
 \includegraphics[width=0.90\linewidth]{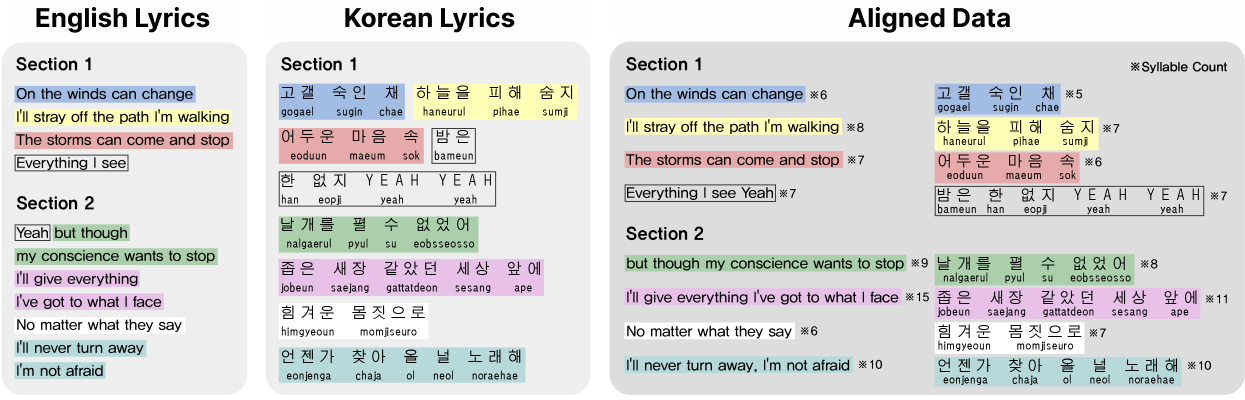}}
\caption{
An example of the alignment task, using ``Beautiful'' by Amber, along with its English and Korean lyrics and their alignments with syllable counts in each language. We obtained the syllable counts by employing the \texttt{syllables} library for English. For Korean, we simply counted the number of characters, as each character in Korean corresponds to one syllable.} 
 \label{fig:alignment}
\end{figure*}

To address these issues, we have compiled a Korean-English lyric translation dataset, of which approximately 89\% comprises lyrics for K-pop songs. This dataset, which contains lyrics for a thousand songs, has been meticulously aligned on a line-by-line and section-by-section basis by humans. The following section of this paper will delve into the construction methodology and details of the dataset. Moving further, we will uncover the unique characteristics of K-pop translation that differentiate it from previously extensively analyzed genres, and demonstrate the application of our dataset to the neural lyric translation task, revealing the necessity of singable lyrics dataset for the enhancement of model performance. We will conclude the paper by highlighting the insights acquired through our experiments using our dataset.

\section{Dataset}

In order to facilitate systematic analysis of lyric translation and advance the development of neural lyric translation models, we introduce a novel singable lyric translation dataset, which comprises pairs of Korean-English lyrics for a thousand songs: 886 K-pop songs, 62 animated musical songs, 34 theatre songs, etc. This dataset incorporates essential metadata such as the name of artist and track, and genre. In addition, it presents meticulous line-by-line and section-by-section alignments of English and Korean lyrics. As we will later show, these alignments, although requiring substantial manual effort as automation is not possible, play a pivotal role in the processes of analysis, neural model development, and evaluation. Although the lyrics cannot be directly downloaded due to copyright issues, they can be accessed via public APIs or URLs. Additionally, we provide an alignment code to aid in both line-wise and section-wise analysis. Our dataset primarily concentrates on K-pop, which constitutes about 89\% of the total data. Nevertheless, we have intentionally included lyrics from other well-studied genres, like animated musical songs (e.g., songs for Disney animation) or theatre songs, to enable comparative analysis across genres.
We believe that this dataset offers value as it reveals insights into K-pop lyric translation, a topic not sufficiently explored in prior research, and the translation of lyrics between Korean and English, languages with substantial grammatical differences. Despite the limited scope of genre and language, we aim to provide insights into the importance of a singable lyric dataset for a diverse array of academic pursuits.
A snippet of the sample data is depicted in Table~\ref{tab:sample}. This dataset is available for download via the provided link~\footnote{\url{https://github.com/havenpersona/lt_dataset}}.

\begin{table}[]
\centering
\resizebox{1.0\linewidth}{!}{%
\begin{tabular}{@{}llll@{}}
\toprule
\textbf{\begin{tabular}[c]{@{}l@{}}Sec-\\tion\\ \#\end{tabular}} & \textbf{\begin{tabular}[c]{@{}l@{}}Line\\ \#\end{tabular}} & \textbf{English} & \textbf{\begin{tabular}[c]{@{}l@{}}Korean\\\textit{(Non-singable English Translation)}\end{tabular}} \\ \midrule

\multirow{4}{*}{1} & 1 & You don't know me & You don't know me \\ \cmidrule(l){2-4} 
 & 2 & L-O-V-E or hatred & L-O-V-E or hatred \\ \cmidrule(l){2-4} 
 & 3 & Hit you with a smile, not goodbye & \begin{tabular}[c]{@{}l@{}}이별 대신 난 순진한 미소만\\ \textit{(Instead of a breakup,} \\ \textit{I only have an innocent smile)}\end{tabular} \\ \cmidrule(l){2-4} 
 & 4 & \begin{tabular}[c]{@{}l@{}}All the while,\\ I'll be sure to leave you wonderin'\end{tabular} & \begin{tabular}[c]{@{}l@{}}오늘도 네 품에 안길래, oh\\ \textit{(So today, I'll embrace you} \\ \textit{in your arms again)}\end{tabular} \\ \midrule
\multirow{4}{*}{2} & 5 & Oh, on the outside I'll be all calm & \begin{tabular}[c]{@{}l@{}}아무것도 모르는 척\\ \textit{(Pretending to not know anything)}\end{tabular} \\ \cmidrule(l){2-4} 
 & 6 & Baby no more real love & Baby, no more real love \\ \cmidrule(l){2-4} 
 & 7 & Imma pretend we're going strong & \begin{tabular}[c]{@{}l@{}}너의 곁에 있어줄게\\ \textit{(I'll stay by your side)}\end{tabular} \\ \cmidrule(l){2-4} 
 & 8 & Then at the end, break your heart & \begin{tabular}[c]{@{}l@{}}마지막엔 break your heart\\ \textit{(In the end, I'll break your heart.)}\end{tabular} \\ \bottomrule
\end{tabular}
}
\caption{Sample data illustrating the English versions and Korean versions of ``Cry for Me'' by Twice}
\label{tab:sample}
\end{table}

\subsection{Source Corpora Collection}
Our dataset includes both pairs of official lyrics for the same songs, that have been officially released in both Korean and English (e.g., a pair of Korean and English versions of ``Cry for Me'' by Twice) and pairs of official Korean lyrics and high-quality unofficial English singable translations (e.g., the official Korean lyrics of ``Attention'' by New Jeans and their singable translation by YouTuber Emily Dimes). To ensure the quality of unofficial translations, we used sources from several YouTubers that we judged to be reliable, after manually reviewing three or more of their translations. Including unofficial translations, which take up 65.2\% of the entire dataset, significantly enhances the size of our dataset.

\subsection{Human Alignment}
Owing to the subjective nature of lyric structure—with no universal agreement on what to call a line and what to call a section—the internet-sourced lyrics for the same song in English and Korean are not aligned on a line-by-line or section-by-section basis. Despite this, the neural model development, evaluation, and analysis of singable lyric translations require these alignments to identify the line-wise and section-wise correspondence and relationship. We observed, however, that automatically generating these alignments is currently unattainable. Some might propose syllable counting as a potential alignment method. However, as it is common to modify melodies to fit varying syllable counts, this method is not practical~\cite{low2008, hui2019translation}. Furthermore, the inclusion of non-lexical vocables, like ``oohs'' and ``aahs,'' in one language's lyrics, and their absence in the other, creates additional inconsistency and challenges in auto-alignment. Consequently, we manually aligned lyrics line-by-line and section-by-section to ensure that lyrics on the same line share the same melodies and sections are divided by the same criteria. To demonstrate this alignment task, we provide Figure~\ref{fig:alignment}, which uses ``Beautiful'' by Amber as an example. This figure illustrates three main points: 1) the different ways sections and lines can be separated, 2) how the same line can consist of a varying number of syllables, and 3) how the same nonlexical sound can be represented in different ways (e.g., ``yeah'' in English lyrics and ``yeah yeah'' in Korean lyrics), which makes the auto-alignment unfeasible. On the other hand, to ensure section-wise alignment, we divided both the English and Korean texts in accordance with the division of the internet-sourced lyrics in the original language. For example, since the original language of ``Beautiful'' by Amber is Korean, we divided the English lyrics into sections by mirroring the structure of the internet-sourced Korean lyrics.


\section{Unpacking K-pop Translation}

In this section, we aim to quantitatively compare the attributes of K-pop translations with those of other extensively researched genres and identify the unique features of K-pop that set its translation process apart. We are basing our comparison on official translations of 234 K-pop songs, 62 animated musical songs, and 34 musical theatre songs in our dataset.

\begin{table}[]
\centering
\resizebox{0.68\linewidth}{!}{%
\begin{tabular}{@{}ccc@{}}
\toprule
\multirow{2}{*}{\textbf{Genre}} & \multicolumn{2}{c}{\textbf{Similarity}} \\ \cmidrule(l){2-3} 
 & $Sem_{line}$ & $Sem_{sec}$ \\ \midrule
K-pop (included) & 0.65 & 0.59 \\
K-pop (excluded) & {\textbf{\underline{0.30}}} & {\textbf{\underline{0.56}}} \\
Animation & 0.45 & 0.60 \\
Theatre & 0.32 & 0.53 \\ \bottomrule
\end{tabular}
}
\caption{Line-wise and section-wise semantic similarity between the original and translated lyrics for K-pop songs, considering both instances where untranslated English lyrics are included and where they are excluded, animated musical songs, and theatre songs.}
\label{tab:semantics}
\end{table}

\subsection{Semantic Pattern} 

A unique characteristic of K-pop lies in its incorporation of both Korean and English within song lyrics. Upon analyzing the K-pop songs in our dataset, we found 30.2\% of the lines are entirely in English, and a blend of English and Korean is observed in 20.7\% of lines. In K-pop song translations into English, English lyrics often remain untranslated, as illustrated in line 1, 2, and 6 of sample data provided in Table~\ref{tab:sample}. As a result, a superficial comparison between K-pop and other genres could lead to the misconception that K-pop has a high line-wise semantic similarity between the original and translated lyrics. However, we observed that this differs from real-world K-pop translation practices, which tend to focus on section-by-section relationships. Using sample data from Table~\ref{tab:sample} as an example, the English lyrics in line 5, ``Oh, on the outside I'll be all calm,'' don't directly align semantically with the Korean lyrics, ``Pretending to not know anything.'' However, when viewed at the section level, the English and Korean lyrics of section 2 show the semantical relatedness by sharing a love theme and a playful mood.

For our analysis, we numerically assessed semantic textual similarity (\textit{sts}) between English and Korean lyrics. In order to do this, we followed a method previously suggested~\cite{ismir2023}: calculating the cosine similarity between the embeddings of the original and translated lyrics, generated by a pre-trained sentence embedding model~\cite{reimer_2019}, \texttt{all-MiniLM-L6-v2}~\cite{minilm2020}. Because this model was trained using English language data, Korean lyrics were automatically translated into English using Google Translator before getting their embeddings. For instance, the semantic textual similarity between $x_i$ = ``하나로 담긴 세상 (a world encapsulated as one)'' and $\Tilde{x_i}$ = ``log in together as one'' ($sts(x_i, \Tilde{x_i})$) would be determined by calculating the cosine similarity between the embeddings of ``a world encapsulated as one'' and ``log in together as one''. For a comparative analysis, we obtained both line-wise and section-wise semantic similarity. Given a pair of lyrics $\mathbf{X}$ and $\Tilde{\mathbf{X}}$, each comprising $m$ sections $\{X_1,...,X_m\}$, $\{\Tilde{X_1},...,\Tilde{X_m}\}$ and $n$ lines $\{x_1, ..., x_n\}$, $\{\Tilde{x_1}, ..., \Tilde{x_n}\}$, previous study defined the line-wise semantic similarity ($Sem_{line}$) and the section-wise semantic similarity ($Sem_{sec}$) as below~\cite{ismir2023}, where $\mathsf{n}(X_i)$ refers to the number of lines in the $i$-th section. (Therefore, $n$ = $\mathsf{n}(X_1)$ + \dots + $\mathsf{n}(X_m)$.)


\begin{equation}
\begin{aligned}
\scalebox{1.00}{$Sem_{line}(\mathbf{X
},\Tilde{\mathbf{X}})=\sum_{i=1}^{n} (\frac{1}{n}sts(x_i, \Tilde{x_i}))$} .
\end{aligned}
\end{equation}
\begin{equation}
\begin{aligned}
\scalebox{0.98}{$Sem_{sec}(\mathbf{X},\Tilde{\mathbf{X}})=\sum_{i=1}^{m} (\frac{\mathsf{n}(X_i)}{n}sts(X_i, \Tilde{X_i}))$} .
\end{aligned}
\label{eq:sim_sem}
\end{equation}

\begin{figure}[t]
 \centerline{
 \includegraphics[width=0.60\linewidth]{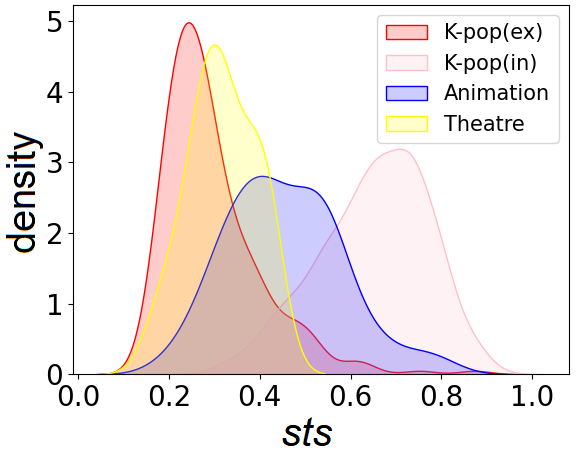}}
\caption{{Density plots showing the distribution of line-by-line semantic textual similarity ($sts$) between the English and Korean lyrics for K-pop songs, considering both instances where untranslated English lyrics are included and where they are excluded, animated musical songs, and theatre songs.}}
 \label{fig:density_plot}
\end{figure}

Table~\ref{tab:semantics} provides the line-wise ($Sem_{line}$) and section-wise ($Sem_{sec}$) semantic similarities between the original and translated lyrics for K-pop (both that include and exclude untranslated English lyrics), animated musical and theatre songs. As expected, K-pop displays the highest line-wise semantic similarity ($Sem_{line}$) when untranslated lyrics are included. However, when those duplicated parts are excluded, this similarity dramatically drops to the lowest level among all the groups. This is contrasted to animated musical songs, which typically need to maintain a strong line-by-line semantic similarity due to their close ties with corresponding visual animations (see Fig~\ref{fig:density_plot}). In contrast, the section-wise semantic similarity ($Sem_{sec}$) stays relatively steady, even with the exclusion of untranslated parts. Therefore, it can be inferred that the semantic relatedness between the original and translated K-pop lyrics exists on a section-by-section basis rather than on a line-by-line basis.

\subsection{Phoneme Repetition Pattern}

The highly repetitive nature of K-pop is reflected not only in melodies but also in the corresponding lyrics, as they are composed to complement music~\cite{nlp_lyrics2005}. To quantify the degree of phoneme repetition, we leveraged the concept of \emph{phoneme distinct-$2$} ($pho$), defined as the ratio of distinct phoneme bigrams to the total number of phoneme bigrams~\cite{ismir2023}.

\begin{equation}
\begin{aligned}
\scalebox{0.95}{$pho(X_i) = \frac{\mbox{\it unique\;bi-gram\;\;in}\;X_i}{\mbox{\it total\;bi-gram\;\;in}\;X_i}$}
\end{aligned}
\label{eq:pho}
\end{equation}

For example, consider a section with two lines $X_1$ = \{``On the winds can change'', ``I'll stray off the path I'm walking''\}. First, the section is decomposed into phonemes and \texttt{\textless{}eos\textgreater} token is added to each line: \lq\lq AA\rq\rq, \lq\lq N\rq\rq, \dots ,\lq\lq CH\rq\rq, \lq\lq EY\rq\rq, \lq\lq N\rq\rq, \lq\lq JH\rq\rq, \lq\lq\texttt{\textless{}eos\textgreater}\rq\rq, \lq\lq AY\rq\rq, ..., \lq\lq W\rq\rq, \lq\lq AO\rq\rq, \lq\lq K\rq\rq, \lq\lq IH\rq\rq, \lq\lq NG\rq\rq, and \lq\lq\texttt{\textless{}eos\textgreater}\rq\rq. Next, the decomposed components are grouped into bi-grams: \lq\lq AAN\rq\rq, ..., \lq\lq NJH\rq\rq,  \lq\lq JH\texttt{\textless{}eos\textgreater}\rq\rq, ...,  \lq\lq IHNG\rq\rq, \lq\lq NG\texttt{\textless{}eos\textgreater}\rq\rq. Finally, the \emph{pho} of the section $X_1$ is obtained by dividing the number of unique bigrams by the total number of bigrams. Given that the number of unique bigrams decreases with the repetition of phonemes, a low \emph{pho} value implies a higher degree of repetition, and a higher ratio suggests the opposite.

\begin{table}[t]
\centering
\resizebox{0.7\linewidth}{!}{%
\begin{tabular}{@{}ccccc@{}}
\toprule
\multirow{2}{*}{\textbf{Genre}} & \multicolumn{2}{c}{\textbf{$Pho_{deg}$}} & \multicolumn{2}{c}{\textbf{$Pho_{var}$}} \\ \cmidrule(l){2-5} 
 & en & kr & en & kr \\ \midrule
K-pop & 0.69 & 0.67 & 0.15 & 0.15 \\
Animation & 0.79 & 0.79 & 0.10 & 0.10 \\ 
Theatre & 0.79 & 0.77 & 0.09 & 0.09 \\ \bottomrule
\end{tabular}
}
\caption{The average values of $Pho_{deg}$ and $Pho_{var}$. Each signifies the degree and variability of phoneme repetition, respectively.}
\label{tab:pho}
\end{table}

\begin{figure*}[!t]
     \centering
    \begin{subfigure}[t]{1.00\textwidth}
        \raisebox{-\height}{\includegraphics[width=\textwidth]{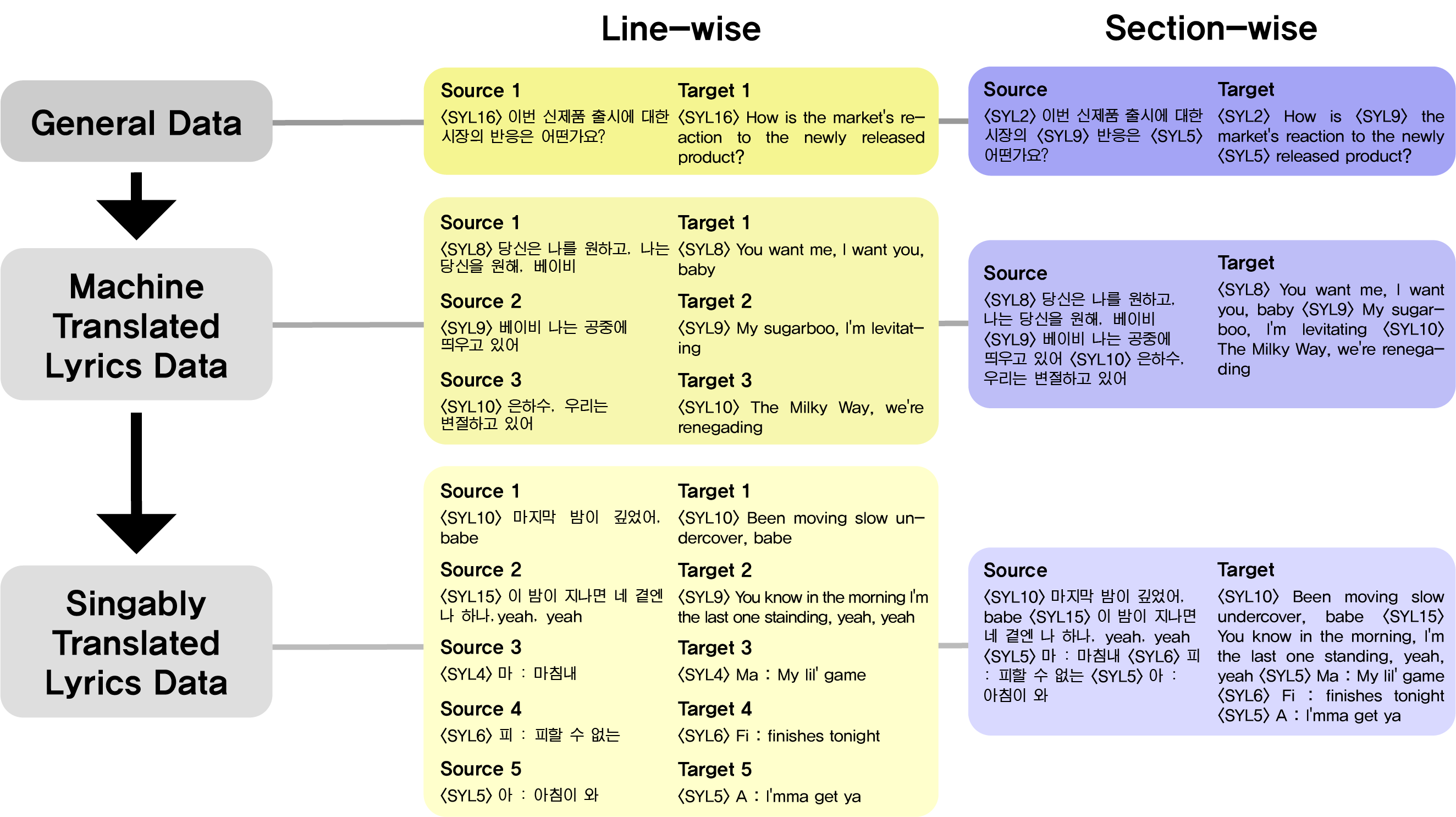}}
    \end{subfigure}
    \caption{Data utilization order during the training phase}
    \label{fig:order}
\end{figure*}

To numerically represent the degree of phoneme repetition for a single song ($Pho_{deg}$), we averaged out the \emph{pho} value across all sections. To capture the variability of phoneme repetition degree for a song ($Pho_{var}$) , we computed the standard deviation of the \emph{pho} values for each section within the song. Table~\ref{tab:pho} presents the average degree and variability ($Pho_{deg}$ and $Pho_{var}$) of phoneme repetition for K-pop, animated musical, and theatre songs in our dataset. K-pop displays the lowest average value, which suggests a high level of repetition inherent to the genre. Additionally, it is noteworthy that K-pop has the highest $Pho_{var}$ value. This denotes significant variability, implying that a typical K-pop song features a mix of highly repetitive sections and others that are less so. As suggested by previous studies, the extent of phoneme repetition in original lyrics is mirrored in translated versions~\cite{drinker1952, low2008}. This is evidenced by the similar $Pho_{deg}$ and $Pho_{var}$ values between the English and Korean versions in each genres.

\section{Neural K-pop Translation}

One of the research opportunities that our dataset provides is the development of a model that can automatically generate singable translations of lyrics, using only textual data. Although past studies have relied on semi-supervised approaches with human-translated non-singable lyrics due to the unavailability of a public singable lyric translation dataset~\cite{acl2022, acl2023}, we present an example of building a neural network model that automatically translates Korean pop lyrics into English, using our dataset. To underscore the potential role of a singable lyric translation dataset, we will further contrast the outcomes of a fully semi-supervised approach versus a fine-tuning approach. The usage examples are provided with two different approaches, line-wise and section-wise. The results of these approaches will be compared with those of a pre-trained English to Koren translation model which isn't specifically designed for lyric translation but shares the same architecture as our models.

\begin{table}[t!]
\centering
\resizebox{0.6\linewidth}{!}{%
\begin{tabular}{@{}ll@{}}
\toprule
\textbf{Hyperparameters} & \textbf{Value} \\ \midrule
\# of Layers & 6 \\
\# of Attention heads & 8 \\
Batch size & 8 \\
Max position embeddings & 1024 \\
Warmup steps & 500 \\
Weight decay & 0.1 \\
Model size & 2048 \\
Vocabulary size & 65051 \\ \bottomrule
\end{tabular}
}
\caption{Model Hyperparameters.}
\label{tab:hyper}
\end{table}

\subsection{Training}
Due to the scarcity of both non-singable and singable lyric translation datasets, a previous study initially trained an English-to-Mandarin lyric translation model with a general Mandarin-English machine translation dataset~\cite{acl2022}. Subsequently, the model was trained using non-singable human-translated lyrics, treating the original lyrics as the target and their translations as the source. In parallel to this previous approach, we began training a transformer-based model~\cite{vaswani2017}, which adopts the architecture of the Marian MT model~\cite{junczys2018marian}, using general translation dataset, with tokenizing source and target lyrics with a pre-trained Korean-English tokenizer~\cite{tiedemann-thottingal-2020-opus}. This was followed by the use of non-singable machine-translated lyrics (instead of non-singable human-translated lyrics, owing to the scarcity of aligned data pairs of English lyrics and human-translated non-singable Korean lyrics). Finally, we fine-tuned the model with our singable lyric translation dataset. Unlike the previous methodology that used melody information along with non-singable lyrics as input, we only used textual data. Key hyperparameters utilized during this process are detailed in Table~\ref{tab:hyper}.

\subsection{Data Preprocessing} 
In order to match the syllable count, the previous study built an English-to-Mandarin lyric translation model that integrated syllable tokens, which represent the number of syllables, at the beginning of each source and target lyric line~\cite{acl2022}. Because the efficacy of these tokens has been only proven in Mandarin text generation, where one character consistently corresponds to one syllable, we compare models with syllable tokens against those without them in order to study the impact of syllable tokens on English text generation, where the count of characters does not necessarily reflect the number of syllables.

We modified data in two different ways: line-wise and section-wise, as seen in Figure~\ref{fig:order}. Below are the details of data modification methods for models that incorporate syllable tokens (\texttt{\textless{}SYL\textgreater}). The methods to construct data for building models without \texttt{\textless{}SYL\textgreater} remain the same except for the omission of the syllable tokens, and that sections are split using \texttt{\textless{}SEP\textgreater} tokens instead of \texttt{\textless{}SYL\textgreater} tokens in the section-wise approach.

\textbf{General}
We obtained 500,000 pairs of Korean sentences and their corresponding English translations \cite{aihub2021}. For line-wise training, we simply incorporated syllable tokens \texttt{\textless{}SYL$s$\textgreater}, where the value of $s$ represents the total syllable count of each target sentence. As an example, ``annyeonghaseyo'' and its English correspondence ``Hello'' would be presented as ``\texttt{\textless{}SYL2\textgreater} annyeonghaseyo'' and ``\texttt{\textless{}SYL2\textgreater} Hello'' because ``Hello'' consists of two syllables. Note that the $s$ value does not have any relation with syllable counts for Korean segments. For section-wise training, we randomly divided both Korean and English sentences into $n$ segments. Given that the syllable counts for each English segment are \{$s_1, ..., s_n$\}, we inserted tokens \{\texttt{\textless{}SYL$s_1$\textgreater}, ..., \texttt{\textless{}SYL$s_n$\textgreater}\} prior to each segment, both in English and Korean, while the order of <SYL$n$> remained the same in both languages.

\textbf{Non-singable Lyrics}
We sourced 10,000 English lyrics randomly from the internet, which were then automatically translated into Korean using Google Translator. Here, the original English lyrics acted as the target, while the machine-translated Korean lyrics were used as the source. For line-wise training, syllable tokens representing the total syllable count of each line were inserted. For section-wise training, we determined the syllable counts of each target line \{$s_1, ..., s_n$\} within a section. Consequently, tokens \{\texttt{\textless{}SYL$s_1$\textgreater}, ..., \texttt{\textless{}SYL$s_n$\textgreater}\} were placed preceding each respective source and target line. The order of lines within a section was randomly shuffled for data augmentation while the shuffled order remained the same in both languages, as we observed that a section with shuffled lines often still lyrically makes sense.

\textbf{Singable Lyrics}
We employed our singable lyric translation dataset to fine-tune our models. For both line-wise and section-wise training, we inserted syllable tokens before each line. In the line-wise approach, each line was treated as individual data, while in the section-wise approach, the entire section was considered one data unit and the order of lines within a section was randomly shuffled.

\subsection{Evaluation Metrics}
Researchers have suggested that traditional methods for analyzing conventional text generation are not appropriate for lyrics, given their unique linguistic characteristics~\cite{lip}. As a result, prior study for automatically assessing lyric translations concentrated on comparing lyrical features of the original lyrics to those of the translated lyrics without using a reference~\cite{ismir2023} rather than using traditional metrics for machine translation evaluation. In alignment with these previously suggested methods, we compared the source lyrics to those automatically translated by our models, focusing on syllable count, semantics, and phonetics

To numerically assess the generated lyrics' syllable count, which is one of the most important factors that determine the singability, we used two metrics: the error rate and the syllable count distance (SCD). The error rate is defined as the rate at which the model generates lines with incorrect syllable counts. On the other hand, the SCD is defined in the following way~\cite{ismir2023}. Suppose that we have a pair of original lyrics $\mathbf{X}$ and translated lyrics $\Tilde{\mathbf{X}}$, each consisting of $n$ lines,  where syllable counts for each line are represented as ${\{s_1, ..., s_n\}}$ and ${\{\Tilde{s_1}, ..., \Tilde{s_n}\}}$. For example, if ``I'll stray off the path I'm walking'' is the second line of the Korean lyrics, and ``haneureul-pihae-sumji (하늘을 피해 숨지)'' is its equivalent line in the English lyrics, then $s_2$ equals 8 and $\Tilde{s_2}$ equals 7. The SCD is calculated as shown below.

\begin{equation}
\begin{aligned}
\scalebox{0.98}{$
{SCD{(\textbf{X},\Tilde{\textbf{X}})}=\frac{1}{2n}\sum_{i=1}^{n} (\frac{|{s_i} - {\Tilde{s_i}}|}{s_i}+} {\frac{|s_i - {\Tilde{s_i}}|}{\Tilde{s_i}}})
$}
\end{aligned}
\end{equation}

As for the semantics, we evaluated section-wise semantic similarity by using Equation~\ref{eq:sim_sem} ($Sem_{sec}$) and semantic coherence between lines by using the BERT-based next sentence prediction (NSP) model~\cite{bert}. While the NSP task was originally proposed to predict whether two sentences given are logically connected, we used the model to evaluate whether two consecutive lines are generated in a coherent manner. To achieve this, we fine-tuned a pre-trained NSP model, \texttt{bert-based-uncased}\footnote{https://huggingface.co/google-bert/bert-base-uncased}, using English lyrics from 7,103 songs that are not included in our training and evaluation data. Given all pairs of consecutively generated (translated) lines, we averaged out the predicted probability of whether two lines are consecutive or not, which we will call the NSP score in this paper.

Finally, we quantitatively assessed the degree and variability of phoneme repetition by obtaining the average value and standard deviation of the \emph{pho} across all sections ($Pho_{deg}$ and $Pho_{var}$) for each song as we did in the previous section.


\subsection{Quantitative Results}

Given that none of our evaluation methods require a reference, we used external data: lyrics from 2,038 K-pop songs that are divided by sections, but not accompanied with corresponding English translations. We ensured that these lyrics did not duplicate any songs present in our training data. When drawing inferences from models with \texttt{\textless{}SYL$s$\textgreater{}} tokens, the value of $s$ during the inference phase was determined based on the syllable count of each line in the source lyrics contrary to the training phase, where the $s$ value was introduced based on the syllable count of the target lyrics. To generate inferences for a single section, composed of $n$ lines, the line-wise approach models inferred on a line-by-line basis, thereby requiring $n$ inference iterations, while the generation of section-wise approach models involved one iteration. The baseline model made inference only on a line-by-line basis, as it does not have the ability to split a translated section into lines. To draw inference, we used the beam search method, one of the most popular search methods in machine translation tasks~\cite{mtsearch2021}, with four beams. 

\begin{table}[]
\centering
\resizebox{0.98\linewidth}{!}{%
\begin{tabular}{@{}cccccc@{}}
\toprule
\multirow{2}{*}{\textbf{\begin{tabular}[c]{@{}c@{}}Input/\\ Output\\ Form\end{tabular}}} & \multirow{2}{*}{\textbf{\begin{tabular}[c]{@{}c@{}}Training\\ Approach\end{tabular}}} & \multicolumn{2}{c}{\textbf{Syllable Count}} & \multicolumn{2}{c}{\textbf{Semantics}} \\ \cmidrule(l){3-6} 
 &  & SCD & \begin{tabular}[c]{@{}c@{}}Error\\ Rate\end{tabular} & $Sem_{sec}$ & NSP \\ \midrule
\multicolumn{2}{c}{\begin{tabular}[c]{@{}c@{}}Baseline\\ ~\cite{junczys2018marian}\end{tabular}} & \begin{tabular}[c]{@{}c@{}}0.34\\ (0.10)\end{tabular} & 0.77 & \begin{tabular}[c]{@{}c@{}}0.71\\ (0.08)\end{tabular} & \begin{tabular}[c]{@{}c@{}}0.47\\ (0.08)\end{tabular} \\ \midrule
\multirow{4}{*}{Line-wise} & Semi-supervised & \begin{tabular}[c]{@{}c@{}}0.29\\ (0.31)\end{tabular} & 0.75 & \begin{tabular}[c]{@{}c@{}}0.64\\ (0.09)\end{tabular} & \begin{tabular}[c]{@{}c@{}}0.51\\ (0.07)\end{tabular} \\ \cmidrule(l){2-6} 
 & \begin{tabular}[c]{@{}c@{}}Semi-supervised \\(+\texttt{\textless{}SYL\textgreater{}})\end{tabular} & \begin{tabular}[c]{@{}c@{}}0.08\\ (0.20)\end{tabular} & 0.48 & \begin{tabular}[c]{@{}c@{}}0.68\\ (0.09)\end{tabular} & \begin{tabular}[c]{@{}c@{}}0.50\\ (0.08)\end{tabular} \\ \cmidrule(l){2-6} 
 & Fine-tuned & \begin{tabular}[c]{@{}c@{}}0.16\\ (0.16)\end{tabular} & 0.69 & \begin{tabular}[c]{@{}c@{}}0.66\\ (0.10)\end{tabular} & \begin{tabular}[c]{@{}c@{}}0.49\\ (0.07)\end{tabular} \\ \cmidrule(l){2-6} 
 & \begin{tabular}[c]{@{}c@{}}Fine-tuned\\(+\texttt{\textless{}SYL\textgreater{}})\end{tabular} & \begin{tabular}[c]{@{}c@{}}0.08\\ (0.38)\end{tabular} & 0.42 & \begin{tabular}[c]{@{}c@{}}0.62\\ (0.12)\end{tabular} & \begin{tabular}[c]{@{}c@{}}0.51\\ (0.07)\end{tabular} \\ \midrule
\multicolumn{1}{l}{\multirow{4}{*}{Section-wise}} & Semi-supervised & \begin{tabular}[c]{@{}c@{}}0.45\\ (0.56)\end{tabular} & 0.78 & \begin{tabular}[c]{@{}c@{}}0.64\\ (0.12)\end{tabular} & \begin{tabular}[c]{@{}c@{}}0.58\\ (0.07)\end{tabular} \\ \cmidrule(l){2-6} 
\multicolumn{1}{l}{} & \begin{tabular}[c]{@{}c@{}}Semi-supervised \\(+\texttt{\textless{}SYL\textgreater{}})\end{tabular} & \begin{tabular}[c]{@{}c@{}}0.18\\ (0.12)\end{tabular} & 0.71 & \begin{tabular}[c]{@{}c@{}}0.68\\ (0.09)\end{tabular} & \begin{tabular}[c]{@{}c@{}}0.54\\ (0.07)\end{tabular} \\ \cmidrule(l){2-6} 
\multicolumn{1}{l}{} & Fine-tuned & \begin{tabular}[c]{@{}c@{}}0.23\\ (0.28)\end{tabular} & 0.73 & \begin{tabular}[c]{@{}c@{}}0.59\\ (0.12)\end{tabular} & \begin{tabular}[c]{@{}c@{}}0.57\\ (0.08)\end{tabular} \\ \cmidrule(l){2-6} 
\multicolumn{1}{l}{} & \begin{tabular}[c]{@{}c@{}}Fine-tuned \\(+\texttt{\textless{}SYL\textgreater{}})\end{tabular} & \begin{tabular}[c]{@{}c@{}}0.09\\ (0.11)\end{tabular} & 0.56 & \begin{tabular}[c]{@{}c@{}}0.60\\ (0.11)\end{tabular} & \begin{tabular}[c]{@{}c@{}}0.54\\ (0.07)\end{tabular} \\ \midrule
\multicolumn{2}{c}{Dataset} & \begin{tabular}[c]{@{}c@{}}0.10\\ (0.04)\end{tabular} & - & \begin{tabular}[c]{@{}c@{}}0.59\\ (0.15)\end{tabular} & \begin{tabular}[c]{@{}c@{}}0.57\\ (0.07)\end{tabular} \\ \bottomrule
\end{tabular}
}
\caption{Comparative evaluation of syllable count and semantics, presenting the average value and associated standard deviation for each metric.}
\label{tab:neural_stats}
\end{table}

\textbf{Syllable Count and Semantics} 
When the models were fine-tuned with our dataset, there was a notable reduction in the average SCD and error rate (see Table~\ref{tab:neural_stats}). For instance, after fine-tuning the section-wise model using our dataset without \texttt{\textless{}SYL\textgreater{}}, the SCD dropped from 0.45 to 0.23, and the error rate decreased from 0.78 to 0.73. This suggests that the model could adapt to match syllable counts when learning from pairs of singable lyrics, even without explicit training for syllable count matching. The ability to match syllable counts was further enhanced when models were trained with \texttt{\textless{}SYL\textgreater{}} tokens, as evidenced by the significantly low average SCD and error rate values compared to those without and the baseline model. This underscores the effectiveness of \texttt{\textless{}SYL\textgreater{}} tokens in matching syllable counts in English textual data. 

These improvements were achieved at the expense of semantic similarity, as suggested by the decline in the $Sem_{sec}$ from the baseline model to semi-supervised models and further decline from the semi-supervised to fine-tuned models. We interpret that they have learned to make a balance between semantics and singability. This interpretation aligns with real-world lyric translation practices where semantics are often sacrificed to enhance singability. This is evidenced by the statistics of the fine-tuned models, especially those trained with \texttt{\textless{}SYL\textgreater{}}, that closely mirror those of K-pop songs in our dataset. Therefore, we conclude that the models most accurately emulate real-world translation patterns when trained with our dataset using \texttt{\textless{}SYL\textgreater{}}. (Note that while they achieved a degree of SCD comparable to that of the dataset, they still struggled to maintain the exact syllable counts, as indicated by significant error rates. This is due to the nature of English, distinct from that of Mandarain, where identical tokens do not consistently equate to the same number of syllables and the same phrases are not always perceived to have an equivalent syllable count.)

While achieving decent performance in terms of syllable count and semantic similarity, the line-wise models displayed low semantic coherence, as indicated by a smaller NSP score than the section-wise models. This is due to the inherent characteristic of the line-wise models that requires making inferences without considering preceding or subsequent lines. It is also noteworthy that the baseline model showed the lowest NSP score assumably because the model failed to coherently capture the lyrical nuances of each line.

\begin{table}[!t]
\centering
\resizebox{0.9\linewidth}{!}{%
\begin{tabular}{@{}cccc@{}}
\toprule
\textbf{\begin{tabular}[c]{@{}c@{}}Input/\\ Output\\ Form\end{tabular}} & \textbf{\begin{tabular}[c]{@{}c@{}}Training\\ Approach\end{tabular}} & \textbf{$Pho_{deg}$} & \textbf{$Pho_{var}$} \\ \midrule
\multicolumn{2}{c}{\begin{tabular}[c]{@{}c@{}}Baseline~\cite{junczys2018marian}\end{tabular}} & 0.70 & 0.14 \\ \midrule
\multirow{4}{*}{Line-wise} & Semi-supervised & 0.70 & 0.14 \\ \cmidrule(l){2-4} 
 & \begin{tabular}[c]{@{}c@{}}Semi-supervised (+\texttt{\textless{}SYL\textgreater{}})\end{tabular} & 0.68 & 0.13 \\ \cmidrule(l){2-4} 
 & Fine-tuned & 0.66 & 0.13 \\ \cmidrule(l){2-4} 
 & \begin{tabular}[c]{@{}c@{}}Fine-tuned (+\texttt{\textless{}SYL\textgreater{}})\end{tabular} & 0.64 & 0.13 \\ \midrule
\multirow{4}{*}{Section-wise} & Semi-supervised & 0.61 & 0.18 \\ \cmidrule(l){2-4} 
 & \begin{tabular}[c]{@{}c@{}}Semi-supervised (+\texttt{\textless{}SYL\textgreater{}})\end{tabular} & 0.66 & 0.14 \\ \cmidrule(l){2-4} 
 & Fine-tuned & 0.53 & 0.18 \\ \cmidrule(l){2-4} 
 & \begin{tabular}[c]{@{}c@{}}Fine-tuned (+\texttt{\textless{}SYL\textgreater{}})\end{tabular} & 0.63 & 0.13 \\ \midrule
\multicolumn{2}{c}{Source} & 0.64 & 0.13 \\ \bottomrule
\end{tabular}
}
\caption{Comparative evaluation of phonetic patterns.}
\label{tab:neural_stats2}
\end{table}

\begin{figure*}[t]
 \centerline{
 \includegraphics[width=1.0\linewidth]{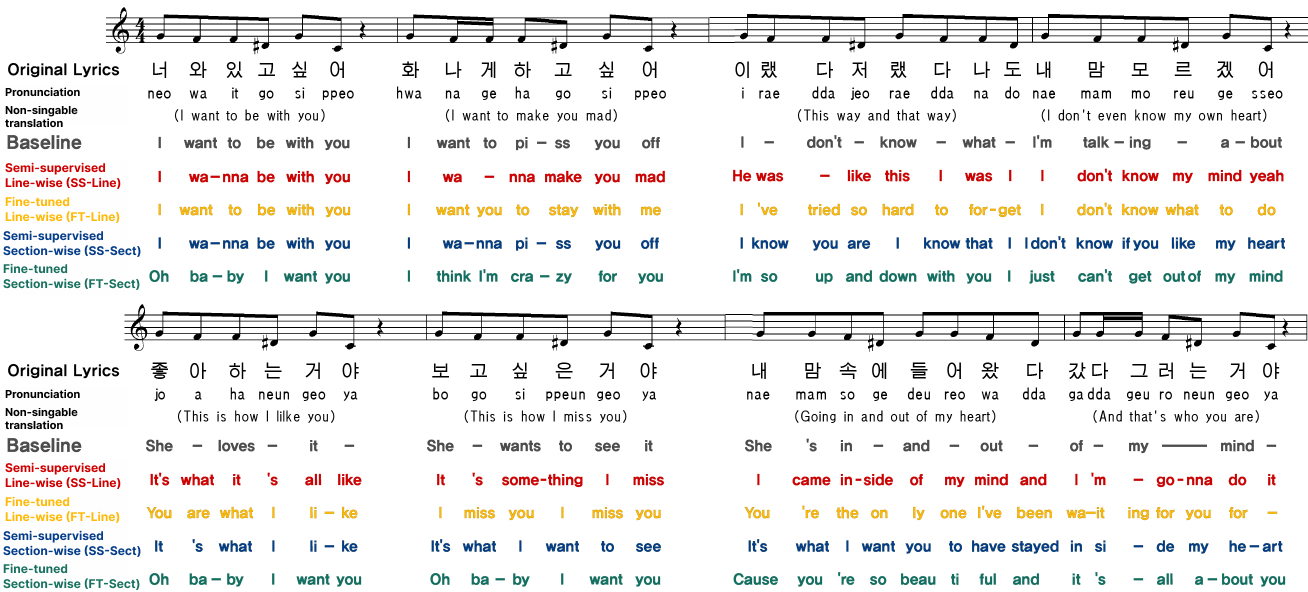}}
\caption{{Automatic translations of ``In \& Out'' by Red Velvet, generated by the baseline model~\cite{junczys2018marian} as well as the semi-supervised and fine-tuned models provided with the original lyrics, their pronunciation, and meanings for comparison. When the syllable count of the generated lyrics exceeds the target count, two or more syllables are put under one note considered to be easily arranged by a music expert. When the syllable count of the generated lyrics is less than the target count, one or more notes, considered ``musically removable'', are not accompanied by lyrics in the score.}}
 \label{fig:inference}
\end{figure*}

\textbf{Phonetic Pattern} 
The baseline model struggles to replicate the repetitive phonetic pattern, an important lyrical characteristic of K-pop, as suggested by a notably higher $Pho_{deg}$ value compared to the source lyrics (see Table~\ref{tab:neural_stats2}). A similar trend is observed in line-wise models without \texttt{\textless{}SYL\textgreater} tokens. On the other hand, the section-wise models without \texttt{\textless{}SYL\textgreater} also displayed significant disparities from the source lyrics in $Pho_{deg}$ value, but in a different manner: these models exhibited excessive repetition with a markedly high degree of variability because they occasionally generated the identical phrase too frequently within a section.

When trained with \texttt{\textless{}SYL\textgreater} tokens, the models adeptly emulated the repetitive nuances of K-pop lyrics with stability, as they know when to continue and when to halt generation. The ability of these models to mirror the repetitive patterns was further enhanced when fine-tuned with our dataset by learning from real-world examples. As a result, they produced English lyrics with $Pho_{deg}$ values aligning closely with the source Korean lyrics in both line-wise and section-wise approaches. Based on our observation that the $Pho_{deg}$ and $Pho_{var}$ values of the original lyrics are reflected in the translated lyrics, we can infer that the model learns the unique phonetic pattern of K-pop when fine-tuned with our dataset using \texttt{\textless{}SYL\textgreater} tokens.

\subsection{Qualitative Results}

We present inference examples drawn by the baseline model as well as self-supervised and fine-tuned models trained with \texttt{\textless{}SYL\textgreater{}} for a section from the K-pop song ``In \& Out'' by Red Velvet in Figure~\ref{fig:inference}. For simplicity, we will denote line-wise and section-wise semi-supervised models as SS-Line and SS-Sect, and line-wise and section-wise fine-tuned models as FT-Line and FT-Sect, respectively.


\textbf{Syllable Counts and Semantics} The baseline model's generated lyrics showed a significant difference in syllable counts with the original lyrics, failing to maintain singability. For instance, the fifth bar contains six notes and therefore, the original lyrics corresponding to this part have six syllables (Jo-a-ha-neun-geo-ya). However, the baseline model produced lyrics with only three syllables, making them unsingable, even with minor melody adjustments. On the other hand, the translations of semi-supervised and fine-tuned models generated lyrics with syllable counts comparable to those of the source lyrics.


The original lyrics in the second bar, ``I want to make you mad (화나게 하고 싶어)'', were translated by the baseline and semi-supervised section-wise model (SS-Sect) into ``I wanna piss you off'', a direct translation of the original meaning. Despite successfully reflecting the original meaning, it contains 6 syllables, while the original lyrics consist of 7 syllables. Conversely, the FT-Sect model successfully generated a 7-syllable line, ``I think I’m crazy for you'', which is not an accurate translation of the corresponding line. However, given that the original lyrics express deep affection for someone, we interpret that the FT-Sect model effectively captured the overall mood and topic of the song, yielding a decent (though not higher than SS-Sect) section-wise semantic similarity ($Sem_{sec}$). Moreover, the output from the FT-Sect model employs more lyrical expressions typically found in English lyrics about love, whereas the SS-Sect model, while achieving literal accuracy, fails to express love naturally in English. A similar tendency is observed in the line-wise models. The semi-supervised line-wise model (SS-line)'s generated lyrics, ``I wanna make you mad'', failed to have accurate syllable counts while achieving semantic accuracy. Conversely, the lyrics translated by the fine-tuned line-wise model (FT-line), ``I want you to stay with me'' maintained the number of syllables but not in semantic accuracy, while successfully capturing the topic, mood, and lyrical expression. These examples further suggest that the models, when fine-tuned with a singable lyrics translation dataset, have learned to prioritize singability over semantic accuracy, reflecting real-world lyrics translation practices.

This example further illustrates the semantic incoherence of line-wise models, particularly the self-supervised model. For example, the consecutive lines, ``He was like this I was I'' and ``I don't know my mind, yeah'', not only lack sensibility but also a logical connection. Conversely, the FT-sect model consistently focuses on expressing love for someone, without performing a direct word-for-word translation. This results in lower semantic accuracy but a reasonably good level of semantic coherence.

\textbf{Phonetic Pattern} The original (source) lyrics and melody lines in Figure~\ref{fig:inference} feature highly repetitive characteristics of K-pop. Similarly, both semi-supervised and fine-tuned models show the repetitive phonetic pattern. However, the ability of the line-wise model to create a sense of repetition is naturally limited to line-wise repetition, as seen in phrases like ``I miss you, I miss you'' in bar 6. Conversely, the section-wise model can generate a sense of repetition on a section-wise basis, as demonstrated in phrases like ``Oh baby I want you'' repeated in bars 1, 5, and 6. This capability results in the FT-sect model having a lower $Pho_{deg}$ value than the FT-line model, as it captures the repeating patterns across the section. 

\section{Conclusions}


In this paper, we introduced a novel singable lyrics dataset that precisely aligns Korean and English lyrics for a thousand songs on a line-by-line and section-by-section basis. As we demonstrated, this alignment is pivotal for analyzing and evaluating lyric translations. Unlike previous translation studies that primarily focused on Western languages and genres, our study targets Korean pop. We utilized this dataset to analyze the unique characteristics of K-pop translations in terms of semantic and phonetic patterns. Additionally, we first suggested that a singable lyrics dataset can be used to build a neural model that translates lyrics into singable forms, even without musical information given, as the model draws inferences from lyrics that are already singable. We compared two approaches to construct a neural lyric translation model, line-wise and section-wise, along with observing the effectiveness of \texttt{\textless{}SYL\textgreater{}} for these approaches, offering insights into the development of neural models capable of translating text akin to lyrics with structured line-by-line and section-by-section characteristics, such as poetry. We hope that this paper will expand the boundaries of singable lyric translation studies and offer valuable insights into this field.

\section{Ethics Statement}
In the pursuit of advancing the field of singable lyric translation, we have considered various ethical aspects to ensure the responsible conduct of our work.
\begin{itemize}
\item \textbf{Transparency:} We are committed to maintaining transparency in our research methodology and dataset creation process. All preprocessing steps, alignment procedures, and model training techniques used in this study are fully disclosed in the paper to enable reproducibility and invite further ethical scrutiny.
\item \textbf{Accessibility:} In contemplation of accessibility to facilitate further research and reproducibility, we will make our dataset publicly available upon acceptancce.
\item \textbf{Inclusivity:} Aiming to contribute to a representation of an under-researched genre, inclusivity and respect for cultural diversity were foundational principles guiding our research.

\end{itemize}
 Finally, we recognize that our research may have broader societal and cultural impacts particularly in promoting cross-cultural understanding through music and language.

\newpage
\nocite{*}
\section{Bibliographical References}\label{sec:reference}

\bibliographystyle{lrec-coling2024-natbib}
\bibliography{lrec-coling2024-example}

\begin{thebibliography}{58}
\expandafter\ifx\csname natexlab\endcsname\relax\def\natexlab#1{#1}\fi

\bibitem[{sin(1954--58)}]{singer-portion-chapter}
 1954--58.
\newblock The late nineteenth century.
\newblock In Charles~Joseph Singer, E.~J. Holmyard, and A.~R. Hall, editors, \emph{A history of technology}, vol.~5. Oxford University Press, London.

\bibitem[{{\AA}kerstr{\"o}m(2010)}]{case_musical2010}
Johanna {\AA}kerstr{\"o}m. 2010.
\newblock Translating song lyrics: a study of the translation of the three musicals by {B}enny {A}ndersson and {B}j{\"o}rn {U}lvaeus.

\bibitem[{Anderman(2017)}]{theatre2017}
Gunilla Anderman. 2017.
\newblock \emph{Europe on stage: Translation and theatre}.
\newblock Bloomsbury Publishing.

\bibitem[{Anon.(1967)}]{anon-67}
Anon. 1967.
\newblock \emph{Title title title title title title title title title title}.
\newblock Organization organization organization.

\bibitem[{BSI(1973{\natexlab{a}})}]{bs-2570-manual}
BSI. 1973{\natexlab{a}}.
\newblock \emph{Natural Fibre Twines}, 3rd edition.
\newblock British Standards Institution, London.
\newblock BS 2570.

\bibitem[{BSI(1973{\natexlab{b}})}]{bs-2570-techreport}
BSI. 1973{\natexlab{b}}.
\newblock Natural fibre twines.
\newblock BS 2570, British Standards Institution, London.
\newblock 3rd. edn.

\bibitem[{Castor and Pollux(1992)}]{CastorPollux-92}
A.~Castor and L.~E. Pollux. 1992.
\newblock The use of user modelling to guide inference and learning.
\newblock \emph{Applied Intelligence}, 2(1):37--53.

\bibitem[{Chercheur(1994)}]{Chercheur-94}
J.L. Chercheur. 1994.
\newblock \emph{Case-Based Reasoning}, 2nd edition.
\newblock Morgan Kaufman Publishers, San Mateo, CA.

\bibitem[{`Chicago'(1982)}]{chicago-82}
`Chicago'. 1982.
\newblock \emph{The {C}hicago Manual of Style}, 13th edition.
\newblock University of Chicago Press.

\bibitem[{Chomsky(1973)}]{chomsky-73}
N.~Chomsky. 1973.
\newblock Conditions on transformations.
\newblock In \emph{A festschrift for {Morris Halle}}, New York. Holt, Rinehart \& Winston.

\bibitem[{Cintr{\~a}o(2009)}]{case2009}
Helo{\'\i}sa~Pezza Cintr{\~a}o. 2009.
\newblock Translating “under the sign of invention”: Gilberto gil’s song lyric translation.
\newblock \emph{Meta}, 54(4):813--832.

\bibitem[{Davies and Bentahila(2008)}]{cultural2008}
Eirlys~E Davies and Abdel{\^a}li Bentahila. 2008.
\newblock Translation and code switching in the lyrics of bilingual popular songs.
\newblock \emph{The Translator}, 14(2):247--272.

\bibitem[{Devlin et~al.(2019)Devlin, Chang, Lee, and Toutanova}]{bert}
Jacob Devlin, Ming-Wei Chang, Kenton Lee, and Kristina Toutanova. 2019.
\newblock \href {https://doi.org/10.18653/v1/N19-1423} {{BERT}: Pre-training of deep bidirectional transformers for language understanding}.
\newblock \emph{Proceedings of the 2019 Conference of the North {A}merican Chapter of the Association for Computational Linguistics (NAACL) : Human Language Technologies, Volume 1}, pages 4171--4186.

\bibitem[{Drinker(1950)}]{drinker1952}
Henry~S. Drinker. 1950.
\newblock On translating vocal texts.
\newblock \emph{The Musical Quarterly}, 36(2):225--240.

\bibitem[{Eco(1990)}]{Eco:1990}
Umberto Eco. 1990.
\newblock \emph{The Limits of Interpretation}.
\newblock Indian University Press.

\bibitem[{Franzon(2005)}]{german_comedy2005}
Johan Franzon. 2005.
\newblock Musical comedy translation: Fidelity and format in the scandinavian my fair lady.
\newblock In \emph{Song and Significance}, pages 263--297. Brill.

\bibitem[{Grandchercheur(1983)}]{Grandchercheur-83}
L.~B. Grandchercheur. 1983.
\newblock \emph{Vers une modélisation cognitive de l'être et du néant}, pages 6--38. Lawrence Erlbaum Associates, Hillsdale, N.J.

\bibitem[{Guo et~al.(2022)Guo, Zhang, Zhang, He, Zhang, Xie, and Boyd-Graber}]{acl2022}
Fenfei Guo, Chen Zhang, Zhirui Zhang, Qixin He, Kejun Zhang, Jun Xie, and Jordan Boyd-Graber. 2022.
\newblock \href {https://doi.org/10.18653/v1/2022.findings-acl.60} {Automatic song translation for tonal languages}.
\newblock In \emph{Findings of the Association for Computational Linguistics (ACL)}, pages 729--743, Dublin, Ireland. Association for Computational Linguistics.

\bibitem[{Hanson(1967)}]{hanson-67}
C.~W. Hanson. 1967.
\newblock Subject inquiries and literature searching.
\newblock In W.~Ashworth, editor, \emph{Handbook of special librarianship and information work}, 3rd edition, pages 414--452.

\bibitem[{Hoel(1971{\natexlab{a}})}]{hoel-71-whole}
Paul~Gerhard Hoel. 1971{\natexlab{a}}.
\newblock \emph{Elementary Statistics}, 3rd edition.
\newblock Wiley series in probability and mathematical statistics. Wiley, New York, Chichester.
\newblock ISBN 0~471~40300.

\bibitem[{Hoel(1971{\natexlab{b}})}]{hoel-71-portion}
Paul~Gerhard Hoel. 1971{\natexlab{b}}.
\newblock \emph{Elementary Statistics}, 3rd edition, Wiley series in probability and mathematical statistics, pages 19--33. Wiley, New York, Chichester.
\newblock ISBN 0~471~40300.

\bibitem[{Howells(1966)}]{howells-66-pop}
W.~W. Howells. 1966.
\newblock Population distances: Biological, linguistic, geographical and environmental.
\newblock \emph{Current Anthropology}, 7:531--540.

\bibitem[{Hui-tung(2019)}]{hui2019translation}
Eos~Cheng Hui-tung. 2019.
\newblock Translation of songs.
\newblock \emph{An Encyclopedia of Practical Translation and Interpreting}, page 351.

\bibitem[{Jespersen(1922)}]{Jespersen:1922}
Otto Jespersen. 1922.
\newblock \emph{Language: Its Nature, Development, and Origin}.
\newblock Allen and Unwin.

\bibitem[{Junczys-Dowmunt et~al.(2018)Junczys-Dowmunt, Grundkiewicz, Dwojak, Hoang, Heafield, Neckermann, Seide, Germann, Aji, Bogoychev, Martins, and Birch}]{junczys2018marian}
Marcin Junczys-Dowmunt, Roman Grundkiewicz, Tomasz Dwojak, Hieu Hoang, Kenneth Heafield, Tom Neckermann, Frank Seide, Ulrich Germann, Alham~Fikri Aji, Nikolay Bogoychev, Andr{\'e} F.~T. Martins, and Alexandra Birch. 2018.
\newblock \href {https://doi.org/10.18653/v1/P18-4020} {{M}arian: Fast neural machine translation in {C}++}.
\newblock In \emph{Proceedings of {ACL} 2018, System Demonstrations}, pages 116--121, Melbourne, Australia. Association for Computational Linguistics.

\bibitem[{Kim et~al.(2023)Kim, Watanabe, Goto, and Nam}]{ismir2023}
Haven Kim, Kento Watanabe, Masataka Goto, and Juhan Nam. 2023.
\newblock A computational evaluation framework for singable lyric translation.
\newblock In \emph{International Society for Music Information Retrieval (ISMIR)}.

\bibitem[{Kingma and Ba(2014)}]{adam_optimizer}
Diederik~P Kingma and Jimmy Ba. 2014.
\newblock Adam: A method for stochastic optimization.
\newblock In \emph{Proceedings of the International Conference on Learning Representations (ICLR)}.

\bibitem[{Leblond et~al.(2021)Leblond, Alayrac, Sifre, Pislar, Jean-Baptiste, Antonoglou, Simonyan, and Vinyals}]{mtsearch2021}
R{\'e}mi Leblond, Jean-Baptiste Alayrac, Laurent Sifre, Miruna Pislar, Lespiau Jean-Baptiste, Ioannis Antonoglou, Karen Simonyan, and Oriol Vinyals. 2021.
\newblock \href {https://doi.org/10.18653/v1/2021.emnlp-main.662} {Machine translation decoding beyond beam search}.
\newblock pages 8410--8434, Online and Punta Cana, Dominican Republic. Association for Computational Linguistics.

\bibitem[{Leni and Pattiwael(2019)}]{disney2019}
Chrisna Leni and Athriyana~Santye Pattiwael. 2019.
\newblock Analyzing translation strategies utilized in the translation of song “{D}o {Y}ou {W}ant to {B}uild a {S}nowman?”.
\newblock \emph{Journal of Language and Literature}, 19(1):55--64.

\bibitem[{Li et~al.(2023)Li, Fan, Bu, Chen, Huang, and Yu}]{li2023translate}
Chengxi Li, Kai Fan, Jiajun Bu, Boxing Chen, Zhongqiang Huang, and Zhi Yu. 2023.
\newblock Translate the beauty in songs: Jointly learning to align melody and translate lyrics.
\newblock \emph{arXiv preprint arXiv:2303.15705}.

\bibitem[{Li et~al.(2016)Li, Galley, Brockett, Gao, and Dolan}]{distinct_n}
Jiwei Li, Michel Galley, Chris Brockett, Jianfeng Gao, and William~B Dolan. 2016.
\newblock A diversity-promoting objective function for neural conversation models.
\newblock In \emph{Proceedings of the 2016 Conference of the North American Chapter of the Association for Computational Linguistics (NACCL) : Human Language Technologies}, pages 110--119.

\bibitem[{Low(2008)}]{low2008}
Peter Low. 2008.
\newblock Translating songs that rhyme.
\newblock \emph{Perspectives: Studies in {T}ranslatology}, 16(1-2):1--20.

\bibitem[{Mahedero et~al.(2005)Mahedero, Mart\'{I}nez, Cano, Koppenberger, and Gouyon}]{nlp_lyrics2005}
Jose P.~G. Mahedero, \'{A}lvaro Mart\'{I}nez, Pedro Cano, Markus Koppenberger, and Fabien Gouyon. 2005.
\newblock \href {https://doi.org/10.1145/1101149.1101255} {Natural language processing of lyrics}.
\newblock In \emph{Proceedings of the 13th Annual ACM International Conference on Multimedia}, pages 475--478.

\bibitem[{Mateo(2012)}]{mateo2012music}
Marta Mateo. 2012.
\newblock Music and translation.
\newblock \emph{Handbook of translation studies}, 3:115--121.

\bibitem[{Medress et~al.(1977)Medress, Cooper, Forgie, Green, Klatt, O'Malley, Neuburg, Newell, Reddy, Ritea et~al.}]{beam1977}
Mark~F. Medress, Franklin~S Cooper, Jim~W. Forgie, CC~Green, Dennis~H. Klatt, Michael~H. O'Malley, Edward~P Neuburg, Allen Newell, DR~Reddy, B~Ritea, et~al. 1977.
\newblock Speech understanding systems: Report of a steering committee.
\newblock \emph{Artificial Intelligence}, 9(3):307--316.

\bibitem[{Ou et~al.(2023)Ou, Ma, Kan, and Wang}]{acl2023}
Longshen Ou, Xichu Ma, Min-Yen Kan, and Ye~Wang. 2023.
\newblock Songs across borders: Singable and controllable neural lyric translation.
\newblock \emph{arXiv preprint arXiv:2305.16816}.

\bibitem[{Park et~al.(2022)Park, Shim, Eo, Lee, Seo, Moon, and Lim}]{aihub2021}
Chanjun Park, Midan Shim, Sugyeong Eo, Seolhwa Lee, Jaehyung Seo, Hyeonseok Moon, and Heuiseok Lim. 2022.
\newblock \href {https://doi.org/10.3390/app12115545} {Empirical analysis of parallel corpora and in-depth analysis using {LIWC}}.
\newblock \emph{Applied Sciences}, 12(11).

\bibitem[{Park and Cho(2014)}]{konlpy2014}
Eunjeong~L. Park and Sungzoon Cho. 2014.
\newblock Ko{NLP}y: Korean natural language processing in {P}ython.
\newblock In \emph{Proceedings of the 26th Annual Conference on Human \& Cognitive Language Technology}, Chuncheon, Korea.

\bibitem[{Reimers and Gurevych(2019)}]{reimer_2019}
Nils Reimers and Iryna Gurevych. 2019.
\newblock \href {https://doi.org/10.18653/v1/D19-1410} {Sentence-{BERT}: Sentence embeddings using {S}iamese {BERT}-networks}.
\newblock In \emph{Proceedings of the 2019 Conference on Empirical Methods in Natural Language Processing and the 9th International Joint Conference on Natural Language Processing (EMNLP-IJCNLP)}, pages 3982--3992.

\bibitem[{Rohma(2018)}]{disney2018}
Irma Aulia~Nur Rohma. 2018.
\newblock \emph{An Analysis of Strategies and Quality of Song Lyric Translation on Disney Movies Soundtracks}.
\newblock Ph.D. thesis, University of Muhammadiyah Malang.

\bibitem[{Singer et~al.(1954--58)Singer, Holmyard, and Hall}]{singer-whole}
Charles~Joseph Singer, E.~J. Holmyard, and A.~R. Hall, editors. 1954--58.
\newblock \emph{A history of technology}.
\newblock Oxford University Press, London.
\newblock 5 vol.

\bibitem[{Snell-Hornby(2007)}]{theatre_opera_2007}
Mary Snell-Hornby. 2007.
\newblock \href {https://doi.org/doi:10.21832/9781853599583-009} {\emph{Theatre and Opera Translation}}, pages 106--119. Multilingual Matters, Bristol, Blue Ridge Summit.

\bibitem[{Spaeth(1915)}]{Spaeth1915}
Sigmund Spaeth. 1915.
\newblock Translating to music.
\newblock \emph{The Musical Quarterly}, 1(2):291--298.

\bibitem[{{Speecon Consortium}(2014)}]{speecon}
{Speecon Consortium}. 2014.
\newblock \emph{Dutch Speecon Database}.
\newblock Speecon Project, distributed via ELRA, Speecon resources, 1.0, ISLRN \href{https://www.islrn.org/resources/613-489-674-355-0}{613-489-674-355-0}.

\bibitem[{Strötgen and Gertz(2012)}]{Martin-90}
Jannik Strötgen and Michael Gertz. 2012.
\newblock Temporal tagging on different domains: Challenges, strategies, and gold standards.
\newblock In \emph{Proceedings of the Eight International Conference on Language Resources and Evaluation (LREC'12)}, pages 3746--3753, Istanbul, Turkey. European Language Resource Association (ELRA).

\bibitem[{Su et~al.(2022)Su, Lan, Wang, Yogatama, Kong, and Collier}]{contrastive2022}
Yixuan Su, Tian Lan, Yan Wang, Dani Yogatama, Lingpeng Kong, and Nigel Collier. 2022.
\newblock \href {https://openreview.net/forum?id=V88BafmH9Pj} {A contrastive framework for neural text generation}.
\newblock In \emph{Advances in Neural Information Processing Systems}.

\bibitem[{Superman et~al.(2000)Superman, Batman, Catwoman, and Spiderman}]{Superman-Batman-Catwoman-Spiderman-00}
S.~Superman, B.~Batman, C.~Catwoman, and S.~Spiderman. 2000.
\newblock \emph{Superheroes experiences with books}, 20th edition.
\newblock The Phantom Editors Associates, Gotham City.

\bibitem[{Susam-Sarajeva(2008)}]{susam2008translation}
{\c{S}}ebnem Susam-Sarajeva. 2008.
\newblock Translation and music: Changing perspectives, frameworks and significance.
\newblock \emph{The Translator}, 14(2):187--200.

\bibitem[{Sutskever et~al.(2014)Sutskever, Vinyals, and Le}]{seq2seq}
Ilya Sutskever, Oriol Vinyals, and Quoc~V. Le. 2014.
\newblock Sequence to sequence learning with neural networks.
\newblock In \emph{Proceedings of the 27th International Conference on Neural Information Processing Systems (NeurIPS)}, page 3104–3112, Cambridge, MA, USA. MIT Press.

\bibitem[{Tiedemann and Thottingal(2020)}]{tiedemann-thottingal-2020-opus}
J{\"o}rg Tiedemann and Santhosh Thottingal. 2020.
\newblock \href {https://aclanthology.org/2020.eamt-1.61} {{OPUS}-{MT} {--} building open translation services for the world}.
\newblock In \emph{Proceedings of the 22nd Annual Conference of the European Association for Machine Translation}, pages 479--480, Lisboa, Portugal. European Association for Machine Translation.

\bibitem[{Vaswani et~al.(2017)Vaswani, Shazeer, Parmar, Uszkoreit, Jones, Gomez, Kaiser, and Polosukhin}]{vaswani2017}
Ashish Vaswani, Noam Shazeer, Niki Parmar, Jakob Uszkoreit, Llion Jones, Aidan~N Gomez, {\L}ukasz Kaiser, and Illia Polosukhin. 2017.
\newblock Attention is all you need.
\newblock In \emph{Proceedings of the Advances in neural information processing systems (NeurIPS)}.

\bibitem[{Wagner(1893)}]{wagner}
Richard Wagner. 1893.
\newblock \emph{Opera and drama, volume II of Richard Wagner’s Prose Works}.
\newblock Bloomsbury Publishing.

\bibitem[{Wang et~al.(2020)Wang, Wei, Dong, Bao, Yang, and Zhou}]{minilm2020}
Wenhui Wang, Furu Wei, Li~Dong, Hangbo Bao, Nan Yang, and Ming Zhou. 2020.
\newblock {MiniLM}: Deep self-attention distillation for task-agnostic compression of pre-trained transformers.
\newblock \emph{Advances in Neural Information Processing Systems (NeurIPS)}, 33:5776--5788.

\bibitem[{Watanabe and Goto(2020{\natexlab{a}})}]{watanabe2020chorus}
Kento Watanabe and Masataka Goto. 2020{\natexlab{a}}.
\newblock A chorus-section detection method for lyrics text.
\newblock In \emph{Proceedings of the 21th International Society for Music Information Retrieval (ISMIR)}, pages 351--359.

\bibitem[{Watanabe and Goto(2020{\natexlab{b}})}]{lip}
Kento Watanabe and Masataka Goto. 2020{\natexlab{b}}.
\newblock Lyrics information processing: Analysis, generation, and applications.
\newblock In \emph{Proceedings of the 1st Workshop on NLP for Music and Audio (NLP4MusA)}, pages 6--12.

\bibitem[{{Winget Ltd.}(1967)}]{winget-67}
{Winget Ltd.} 1967.
\newblock Detachable bulldozer attachment for dumper vehicles.
\newblock GB Patent Specification 1060631.

\bibitem[{Wright(1963)}]{wright-63}
R.~C. Wright. 1963.
\newblock \emph{Report Literature}, pages 46--59.

\bibitem[{Zavatta(1992)}]{Zavatta-92}
A.~Zavatta. 1992.
\newblock \emph{Un Générateur d'Insultes s'intégrant dans un Système de Dialogue Humain-Machine.}
\newblock Thèse de doctorat en informatique, Université Paris-sud, Centre d'Orsay.

\end{thebibliography}

\bigbreak

\newpage

\appendix

\newpage

\begin{table*}[]
\centering
\resizebox{0.7\linewidth}{!}{%
\begin{tabular}{|l|l|l|l|}
\hline
\textbf{\begin{tabular}[c]{@{}l@{}}Sec-\\ tion\\ \#\end{tabular}} & \textbf{\begin{tabular}[c]{@{}l@{}}Line\\ \#\end{tabular}} & \textbf{English (EN)} & \textbf{Korean  (KR)} \\ \hline
\multirow{4}{*}{1} & 1 & I know & I know \\ \cline{2-4} 
 & 2 & Our love ain't anything to fight for & 고쳐 쓸 가치도 없단 걸 \\ \cline{2-4} 
 & 3 & But I'll never break out of the cycle & 하지만 그녀와 달리 난 널 \\ \cline{2-4} 
 & 4 & \begin{tabular}[c]{@{}l@{}}I don't really wanna let you go \\ (Never let go)\end{tabular} & 쉽게 놔줄 맘이 없거든 (Never let go) \\ \hline
\multirow{4}{*}{2} & 5 & You don't know me & You don't know me \\ \cline{2-4} 
 & 6 & L-O-V-E or hatred & L-O-V-E or hatred \\ \cline{2-4} 
 & 7 & Hit you with a smile, not goodbye & 이별 대신 난 순진한 미소만 \\ \cline{2-4} 
 & 8 & \begin{tabular}[c]{@{}l@{}}All the while, I'll be sure to leave you \\ wonderin'\end{tabular} & 오늘도 네 품에 안길래, oh \\ \hline
\multirow{4}{*}{3} & 9 & Oh, on the outside I'll be all calm & 아무것도 모르는 척 \\ \cline{2-4} 
 & 10 & Baby no more real love & Baby, no more real love \\ \cline{2-4} 
 & 11 & Imma pretend we're going strong & 너의 곁에 있어줄게 \\ \cline{2-4} 
 & 12 & Then at the end, break your heart & 마지막엔 break your heart \\ \hline
\multirow{3}{*}{4} & 13 & Bad boy, bad boy & Bad boy, bad boy \\ \cline{2-4} 
 & 14 & \begin{tabular}[c]{@{}l@{}}Yeah, you really make me a mad \\ girl, mad girl\end{tabular} & \begin{tabular}[c]{@{}l@{}}Yeah, you really make me a mad \\ girl, mad girl\end{tabular} \\ \cline{2-4} 
 & 15 & Woah-oh-oh & Woah-oh-oh \\ \hline
\multirow{4}{*}{5} & 16 & I want you to cry, cry for me & I want you to cry, cry for me \\ \cline{2-4} 
 & 17 & \begin{tabular}[c]{@{}l@{}}Thе way I cried for you, baby, cry \\ for me\end{tabular} & 내가 울었던 것처럼 cry for me \\ \cline{2-4} 
 & 18 & Make your rain fall, cry for mе & Make your rain fall, cry for me \\ \cline{2-4} 
 & 19 & But again & But, again \\ \hline
\multirow{5}{*}{6} & 20 & \begin{tabular}[c]{@{}l@{}}Somehow you keep me goin' round \\ and round\end{tabular} & 조금씩 조금씩 또 빠져가 \\ \cline{2-4} 
 & 21 & \begin{tabular}[c]{@{}l@{}}All the walls I built around me come \\ crashin' down\end{tabular} & 사랑에 내 결심이 또 무너져가 \\ \cline{2-4} 
 & 22 & Makin' excuses, gotta drown 'em out & 용서할 핑계를 만들어가 \\ \cline{2-4} 
 & 23 & \begin{tabular}[c]{@{}l@{}}I want you to, I want you to, I want \\ you to cry\end{tabular} & \begin{tabular}[c]{@{}l@{}}I want you to, I want you to, I want \\ you to cry for me\end{tabular} \\ \cline{2-4} 
 & 24 & Hmm, yeah & Hmm, yeah \\ \hline
\multirow{4}{*}{7} & 25 & I don't know if I'm just & I don't know 너란 놈 \\ \cline{2-4} 
 & 26 & In too deep and I'm confused & 미워질 줄 모르고 \\ \cline{2-4} 
 & 27 & \begin{tabular}[c]{@{}l@{}}All my friends hate your guts but I'm \\ still defending you (Ooh; Yah yah, \\ yah yah)\end{tabular} & \begin{tabular}[c]{@{}l@{}}친구들한텐 또 너를 감싸주는 중 \\ (Ooh; Yah yah, yah yah)\end{tabular} \\ \cline{2-4} 
 & 28 & \begin{tabular}[c]{@{}l@{}}I can't seem to cut you loose (Yah \\ yah, yah yah; Mmm, yeah)\end{tabular} & \begin{tabular}[c]{@{}l@{}}바보가 돼 버렸군 (Yah yah, yah \\ yah; Mmm, yeah)\end{tabular} \\ \hline
\multirow{3}{*}{8} & 29 & \begin{tabular}[c]{@{}l@{}}Ooh, don't know how you keep on \\ laughin' everyday\end{tabular} & Ooh, 너 왜 자꾸 나를 보며 웃는데 \\ \cline{2-4} 
 & 30 & \begin{tabular}[c]{@{}l@{}}Just a single tear from you, I'd be \\ okay (Ooh)\end{tabular} & 딱 한 번의 눈물이면 되는데 (Ooh) \\ \cline{2-4} 
 & 31 & \begin{tabular}[c]{@{}l@{}}Cry for me, let me please forgive \\ you, oh\end{tabular} & \begin{tabular}[c]{@{}l@{}}Cry for me, let me please forgive \\ you, oh\end{tabular} \\ \hline
\multirow{4}{*}{9} & 32 & Oh on the outside I'll be all calm & 아무것도 모르는 척 \\ \cline{2-4} 
 & 33 & Baby just like real love & Baby, just like real love \\ \cline{2-4} 
 & 34 & Tellin' myself we're going strong & 마지막 기회야 어서 \\ \cline{2-4} 
 & 35 & If it's the end, break your heart & 보여줘봐 your true love \\ \hline
\multirow{3}{*}{10} & 36 & Bad boy, bad boy & Bad boy, bad boy \\ \cline{2-4} 
 & 37 & \begin{tabular}[c]{@{}l@{}}Yeah, you really make me a sad girl, \\ sad girl (Sad girl, sad girl)\end{tabular} & \begin{tabular}[c]{@{}l@{}}Yeah, you really make me a sad girl, \\ sad girl (Sad girl, sad girl)\end{tabular} \\ \cline{2-4} 
 & 38 & Woah-oh-oh & Woah-oh-oh \\ \hline
\multirow{4}{*}{11} & 39 & I want you to cry, cry for me & I want you to cry, cry for mе \\ \cline{2-4} 
 & 40 & \begin{tabular}[c]{@{}l@{}}Thе way I cried for you, baby, cry \\ for me\end{tabular} & 내가 울었던 것처럼 cry for me \\ \cline{2-4} 
 & 41 & Make your rain fall, cry for mе & Make your rain fall, cry for mе \\ \cline{2-4} 
 & 42 & But again & But again \\ \hline
\multirow{4}{*}{12} & 43 & \begin{tabular}[c]{@{}l@{}}Somehow you keep me goin' round \\ and round\end{tabular} & 조금씩 조금씩 또 빠져가 \\ \cline{2-4} 
 & 44 & \begin{tabular}[c]{@{}l@{}}All the walls I built around me come \\ crashin' down\end{tabular} & 사랑에 내 결심이 또 무너져가 \\ \cline{2-4} 
 & 45 & Makin' excuses, gotta drown 'em out & 용서할 핑계를 만들어가 \\ \cline{2-4} 
 & 46 & \begin{tabular}[c]{@{}l@{}}I want you to, I want you to, I want \\ you to cry for me\end{tabular} & \begin{tabular}[c]{@{}l@{}}I want you to, I want you to, I want \\ you to cry for me\end{tabular} \\ \hline
\end{tabular}
}
\end{table*}

\begin{table*}[t!]
\centering
\resizebox{0.7\linewidth}{!}{%
\begin{tabular}{|l|l|l|l|}
\hline
\multirow{6}{*}{13} & 47 & If love is a game & 사랑이란 게 \\ \cline{2-4} 
 & 48 & Don't want to play & 너무 혹독해 \\ \cline{2-4} 
 & 49 & You poison my veins & 미운 마음도 \\ \cline{2-4} 
 & 50 & Then take it all away & 다 녹아버리게 해 \\ \cline{2-4} 
 & 51 & I'm chasin' that taste & 또 다시 원해 \\ \cline{2-4} 
 & 52 & I want your kiss, yeah, yeah, yeah & 널 내 곁에 yeah, yeah, yeah \\ \hline
\multirow{4}{*}{14} & 53 & I want you to cry, cry for me & I want you to cry, cry for me \\ \cline{2-4} 
 & 54 & \begin{tabular}[c]{@{}l@{}}Can you at least pretend, baby, cry \\ for me\end{tabular} & 너 연기라도 해 빨리 cry for me \\ \cline{2-4} 
 & 55 & Make your rain fall & Make your rain fall \\ \cline{2-4} 
 & 56 & Fall and fall now, yeah & Fall and fall now, yeah \\ \hline
\multirow{4}{*}{15} & 57 & I want you to cry, cry for me & I want you to cry, cry for me \\ \cline{2-4} 
 & 58 & \begin{tabular}[c]{@{}l@{}}The way I cried for you, baby, cry \\ for me\end{tabular} & 내가 울었던 것처럼 cry for me \\ \cline{2-4} 
 & 59 & Make your rain fall, cry for me & Make your rain fall, cry for me \\ \cline{2-4} 
 & 60 & But again & But again \\ \hline
\multirow{4}{*}{16} & 61 & \begin{tabular}[c]{@{}l@{}}Somehow you keep me goin' round \\ and round\end{tabular} & 조금씩 조금씩 또 빠져가 \\ \cline{2-4} 
 & 62 & \begin{tabular}[c]{@{}l@{}}All the walls I built around me come \\ crashin' down\end{tabular} & 사랑에 내 결심이 또 무너져가 \\ \cline{2-4} 
 & 63 & Makin' excuses, gotta drown 'em out & 용서할 핑계를 만들어가 \\ \cline{2-4} 
 & 64 & \begin{tabular}[c]{@{}l@{}}I want you to, I want you to, I want \\ you to die for me\end{tabular} & \begin{tabular}[c]{@{}l@{}}I want you to, I want you to, I want \\ you to die for me\end{tabular} \\ \hline
\end{tabular}
}
\caption{A pair of lyrics for official English and Korean versions of ``Cry for Me'' by Twice}
\label{tab:appendix_sample}
\end{table*}

\clearpage

\section{Sample Data}
\label{sec:sample}
Table~\ref{tab:appendix_sample} shows a pair of official English and Korean versions of ``Cry for Me'' by Twice. Please note that we do not own the rights to these lyrics. Our dataset simply provides public API access to the lyrics and the code for their line-by-line and section-by-section alignment.

\begin{table}[t!]
\centering
\resizebox{0.65\linewidth}{!}{%
\begin{tabular}{@{}ll@{}}
\toprule
\textbf{Property} & \textbf{Count} \\ \midrule
Songs & 1000 \\
Total sections & 11330 \\
Unique sections (Korean) & 9119 \\
Unique sections (English) & 9040 \\
Total lines & 59054 \\
Unique lines (Korean) & 39134 \\
Unique lines (English) & 38536 \\
Unique vocabulary (Korean) & 7570 \\
Unique vocabulary (English) & 16086 \\ \bottomrule
\end{tabular}
}
\caption{Dataset Statistics.}
\label{tab:count}
\end{table}

\section{Dataset Statistics}
\label{sec:statistics}

Table~\ref{tab:count} shows the key properties of our dataset, along with their respective counts. To determine the number of Korean vocabulary items during the statistical gathering process, we used a morphological analyzer~\cite{konlpy2014} with excluding the English parts in the Korean version lyrics. On the other hand, counting the number of English vocabulary was achieved through simple whitespacing. Owing to the repetitive nature of lyrics, the numbers of unique sections and lines are less than the total count of sections and lines.

\section{Training Details}
\label{sec:training}

Each dataset (General, Machine Translated Lyrics, Singably Translated Lyrics) was split in a 9:1 ratio for the training and validation sets. We trained our models to minimize the Negative Log Likelihood (NLL) by comparing the predicted probability distribution to the ground truth at each epoch. We stopped the training if there was no reduction in the loss function observed for three consecutive epochs when tested on the validation set, with optimization achieved through the Adam optimizer~\cite{adam_optimizer}. Key hyperparameters utilized during this process are detailed in Table~\ref{tab:hyper}.

\clearpage

\end{document}